# Modeling Progress in AI

Miles Brundage

Consortium for Science, Policy, and Outcomes, Arizona State University
PO Box 875603, Tempe, AZ 85287
miles.brundage@asu.edu

**Abstract**

Participants in recent discussions of AI-related issues ranging from intelligence explosion to technological unemployment have made diverse claims about the nature, pace, and drivers of progress in AI. However, these theories are rarely specified in enough detail to enable systematic evaluation of their assumptions or to extrapolate progress quantitatively, as is often done with some success in other technological domains. After reviewing relevant literatures and justifying the need for more rigorous modeling of AI progress, this paper contributes to that research program by suggesting ways to account for the relationship between hardware speed increases and algorithmic improvements in AI, the role of human inputs in enabling AI capabilities, and the relationships between different sub-fields of AI. It then outlines ways of tailoring AI progress models to generate insights on the specific issue of technological unemployment, and outlines future directions for research on AI progress.

## Introduction

Recent discussions of AI and its social consequences have emphasized the progress made in AI research thus far, and the additional progress that can be anticipated in the future (Dietterich and Horvitz, 2015; Future of Life Institute, 2015). Yet when it comes to characterizing what exactly progress in AI consists of, what has driven it, and how quickly it might proceed in the future, there is little formal analysis in the literature and apparently much disagreement. For example, surveys of AI researchers on the future of the field have found a wide range of opinions about how long it may take for human-level AI to be achieved (Grace and Cristiano, 2015). Some respondents have also taken issue with the very concept of human-level AI (Kruel 2011), variations of which are often used as a reference point in such surveys. Participants in debates on issues ranging from intelligence explosion to technological unemployment have invoked differing implicit models of AI progress to support their opinions (Brundage, 2015), e.g. in discussing the plausibility of an intelligence explosion or which jobs are most resistant to automation. In light of such diversity of opinions on policy-relevant questions, it seems prudent to methodically critique and improve such models. Well-developed and appropriately caveated models of AI progress could be used, as they already are in other technological domains (Roper et al., 2011), to support the development of plausible future scenarios, and thus to aid in long-term policy analysis and robust decision-making (Lempert et al., 2003).

This paper aims to help address these ambiguities in the literature as follows. First, in the section "Analysis of Relevant Literatures," we discuss several literatures that, while falling short of developing full models of AI progress, help orient such research in fruitful directions. Second, in "Toward Rigorous Modeling of AI Progress," we take some steps toward a rigorous model of AI progress by analyzing issues such as the relationship between hardware and software developments, the role of human inputs in enabling AI capabilities, and the relationships between different sub-fields of AI. Next, in "AI Progress and the Future of Work," we analyze the differing accounts of AI capabilities invoked in the literature on technological unemployment and suggest ways that rigorous AI progress modeling might advance such debates. Finally, "Future Directions" outlines areas for future research on AI progress, and "Conclusion" summarizes the contributions of the paper.

## Analysis of Relevant Literatures

While there is not yet a robust literature theorizing or empirically evaluating the state of the art in AI (Hernandez-Orallo, 2014), let alone extrapolating it, many literatures are relevant to the AI progress modeling research program proposed in this paper. In this section, we detail some of the findings of such literatures, how they help us to think about AI progress, and what their limitations are.

## AI Evaluation Literature

First, as summarized by Hernandez-Orallo (2014), many different methods have been proposed to evaluate the intelligence of AI-based artifacts—what we can call the "evaluation" literature. Hernandez-Orallo distinguishes between "task" evaluations and "ability" evaluations, with the former evaluating agent performance on a relatively specific task (e.g. a specific vision benchmark) and the latter measuring agent performance across a wider range of tasks (e.g. vision or perception in general), although he notes that there is in fact continuity between these levels of analysis. Of particular relevance for the present discussion, Hernandez-Orallo notes an important analogy that we will take up in the next section, namely between the "factors" used to disaggregate intelligence in the human psychometrics literature (e.g. in Carroll, 1993) such as reading and writing ability and visual processing, and the distinctions between sub-fields of AI, such as natural language processing and computer vision. While there is not likely to be a perfect mapping between human intelligence factors and the topology of the AI field, and human psychometrics has different motivations than does AI progress modeling, Hernandez-Orallo's insight usefully suggests the potential utility of a hierarchical model of AI progress. Such a model, further elements of which will be proposed in the next section, would situate general intelligence evaluation at the top of the model, intermediate abilities/fields in the middle, and task level evaluation of agents (or of the achievements of the field as a whole) at the bottom.

In recent years, there has been much discussion of moving "beyond the Turing Test" (You, 2015), and new benchmarks for AI evaluation have been put forward such as the "Visual Turing test" (Geman et al., 2014). There are also discussions of related issues in the context of specific sub-fields of AI, such as the challenges associated with evaluating integrated cognitive systems (Jones et al., 2012; Gomila and Müller, 2012). Beyond these various proposals, there is also methodological literature on the empirical analysis of AI systems (Cohen 1995), and AI textbooks (e.g. Russell and Norvig, 2009; Poole and Mackworth, 2010) have specified dimensions of complexity for agents and environments such as partial versus full observability, single versus multi-agent, etc. that could be incorporated into a model of AI progress.

While these literatures are valuable and point to various insights and pitfalls that AI progress modelers should be aware of, they generally do not constitute full-blown efforts at modeling AI progress (in the sense used here) for two reasons. First, they are static, seeking to benchmark AI progress in the light of a particular agent or capability, without explicitly seeking to represent the dynamics of AI progress (i.e. how we got to the present point, and extrapolating how progress might continue in the future). A dynamic model, taking into account time or some measure of effort applied toward making progress in AI, as has been developed in other domains (Nagy et al., 2013), would be necessary to extrapolate future scenarios in a principled way. Second, the evaluation literature is generally oriented toward developing specific benchmarks of progress and driving additional progress in that area, rather than theorizing the relationships between different benchmarks and the societal drivers of progress on them. Instead of seeking a single benchmark for AI progress, the approach suggested here is aimed at explaining how different benchmarks, at different levels of analysis, fit together, as well as causally explaining progress dynamics, rather than merely accelerating progress.

## Technology Forecasting Literature

Beyond the subject of evaluation, there is also a literature on AI progress prediction in particular and on technological forecasting in general (referred to collectively here as the "forecasting" literature). Grace and Cristiano (2015) find at least 9 surveys about future AI progress. Armstrong and Sotala (2012), Armstrong et al. (2014), and Bostrom (2014) analyze prior predictions of future AI progress. Notably, such efforts at prediction typically focus on a particular benchmark, "human-level AI" (or variations thereof). The problems associated with such a focus (e.g. that AI already vastly exceeds humans in some areas), are discussed by, among others, some of the respondents to Kruel's (2011) survey. For purposes such as anticipating technological unemployment, it is important to anticipate AI progress that falls short of, exceeds, or is orthogonal to "human-level" AI. This suggests the need for a richer account of AI progress than the "time until human-level AI" prediction paradigm. Additionally, prediction is not the only reason to develop models of technological progress—others include encouraging clarity about and scrutiny of underlying assumptions, stimulating creative thinking about the future, and making better decisions in the present (Roper et al., 2011). A model outlining a plausible sequence in which tasks may become easy to automate, for example, might be beneficial for the public and policymakers, even if there are not specific temporal predictions associated with it.

Several findings from the forecasting literature are particularly relevant in thinking about how to model AI progress. First, short-term technology forecasts generally fare better than long-term ones (Mullins, 2012). Second, quantitative technology forecasts generally fare better than qualitative ones (Mullins, 2012). Third, model-based predictions in the domain of AI in particular generally fare better than intuition-based ones (Armstrong et al., 2014). Fourth, there is a significant amount of commonality in the rates of progress across a wide range of technologies, which can often be surprisingly well captured by a simple exponential trend (Nagy et al., 2013); however, determining the right metrics for progress and the independent variables driving that progress is not always easy, and the appropriate parameters for such models need to be investigated empirically—a subject to be revisited in the following section.

**AI Risk Literature**

Next, there is a growing literature on long-term risks associated with AI (Bostrom, 2014). In the service of estimating the nature and magnitude of future risks from advanced AI systems, various researchers (e.g. Bostrom, 2014; Yudkowsky, 2013; Yampolskiy, 2015) have analyzed the potential for AI systems to rapidly grow in intelligence. The concepts and tools used by such researchers are relevant to at least some types of AI progress modeling—for example, Bostrom (2014) develops formulas relating the recalcitrance of intelligence improvements and the optimization power applied to improving intelligence to AI progress, which could be extended in future work. However, these ideas are generally more applicable for the purposes for which they were developed (analyzing future catastrophic risks) than uses such as anticipating the economic consequences of AI, because they often begin from the premise that an agent has human-level (or beyond) capabilities in certain domains. The ideas for progress modeling discussed in this paper don't necessarily presuppose that level of capability.

**Natural Intelligence Literature**

In addition to the literatures on AI evaluation, forecasting, and risk, there is a rich literature on natural intelligence that may be fruitful in developing models of AI progress. For example, work in evolutionary and comparative psychology (Wynne and Udell, 2013) differentiates between types and levels of intelligence possessed on average by different species, possibly suggesting relevant metrics for AI progress modeling. Developmental psychology (Miller, 2009) addresses the development of (among other human characteristics) intelligence across the lifespan, and its theories of intellectual development may suggest insights regarding the recalcitrance of different problems in AI. Finally, cognitive psychology (Anderson, 2009) analyzes the decision-making dimensions of human psychology, and the narrower field of psychometrics provides tools such as IQ tests (and analyses thereof), which have sometimes been applied to evaluating AI systems (Ohlsson et al., 2015).

**Technology Roadmapping Literature**

Finally, there is a literature on technological roadmapping methodologies (Roper et al., 2011) that may assist in the development of AI progress models, and there are developed roadmaps for the future of related fields such as robotics (e.g. Robotics VO, 2013). Such roadmaps are often created by militaries (e.g. US Department of Defense, 2013) to aid in long-term analysis. However, such roadmaps rarely feature detailed methodological information, and often seem to depend on intuition as the basis for assigning, e.g. 5, 10, and 15 year-out milestones for technical progress (these milestones are also typically qualitative). Additionally, different roadmaps use different metrics for progress, and divide up fields like robotics into different sub-fields, without using the sort of principled approach suggested in this paper and discussed in more detail in the next section. Also relevant for our analysis here is the fact that, to the author's knowledge, there is no detailed, explicit roadmap available yet for AI (as opposed to robotics) progress, although the two are related.

## Toward Rigorous Modeling of AI Progress

In light of the limitations of prior work, this section suggests ways to build on the aforementioned literatures and more rigorously model AI progress. In particular, it begins with an exploration of the different levels and units of analysis that might be considered in such modeling; explores the relationship between hardware and software progress; analyzes the human contribution to AI performance from the perspective of AI progress modeling; and suggests ways to account for the diversity of sub-fields of AI and the problem of developing integrated AI systems.

**Levels and Units of Analysis**

AI progress could be modeled at various levels – e.g. at the level of a particular exemplar agent's performance on a wide range of tasks, the conceptual progress demonstrated in the literature, the performance of human-machine systems, or the performance of a range of different agents on different tasks to which they're specialized.

The purpose of a particular exercise in progress modeling is critical in determining the appropriate level of analysis. For example, if one is trying to determine whether a customer service worker's job can be plausibly automated or not, it's important to consider whether certain tasks/abilities can be integrated easily in a single agent—the fact that, e.g. one system demonstrates impressive natural language processing and another demonstrates impressive perception does not imply that these capabilities can be integrated in order to automate the perceptual and linguistic aspects of the worker's job that involves both perception and language. Thus, in such a context (considered in more detail in the section on the future of work below), the ease of integration of cognitive abilities must itself be modeled in some fashion.

In addition to discerning an appropriate level of analysis for AI progress modeling, there is also the question of the appropriate unit of analysis at that level. For example, if we are interested in the state of the art in machine vision, it makes a difference whether our unit of analysis is the performance of a given algorithm using an average consumer computer's hardware, or its performance on the world's largest computer cluster. The former unit of analysis may be more relevant for certain purposes even if the performance of the latter is higher. Additionally, considerations like the availability of data for machine learning in a certain domain, or the extent of human input provided to fine-tune an algorithm's performance, may influence one's appraisal of the state of the art for a given task or ability. Therefore, in developing a general framework for AI progress modeling, it may be more useful to think in terms of "states of the art" rather than a singular "state of the art," and to explicitly state (and if possible, quantify) the various independent variables determining agent performance.

**Hardware and Software Progress**

Many observers have noted the relevance of computer hardware improvements to recent progress in AI, and in particular areas such as deep learning (LeCun et al., 2015; Schmidhuber, 2015). Indeed, exponential improvements in the computing power available for a given price, and the development of particular technologies such as graphics processing units (GPUs), have accelerated progress in multiple AI domains. But what exactly is the relationship between hardware and software progress, and can it be quantified in a way that lends itself to rigorous modeling and extrapolation? This section qualitatively explains several ways in which hardware and software progress interact in order to pave the way for such modeling efforts.

First, improvements in hardware speed enable increases in the level of performance attainable for a given algorithm in many cases. For example, in the case of deep learning, faster CPUs and GPUs have enabled increases in the number of parameters of neural networks and in the size of data sets that those networks can be trained on in a reasonable amount of time, allowing more difficult tasks to be solved. Grace (2013) analyzed several domains and found that algorithmic progress contributed about 50-100% as much to improvement in performance as did hardware progress.

Second, software innovations can enable the efficient use of larger amounts of hardware through parallelization (which, in turn, can also increase the performance of AI systems). For example, DistBelief (Dean et al., 2012) enables the use of large computing clusters for deep learning, and the Gorila architecture (Nair et al., 2015) does the same for deep reinforcement learning (Mnih et al., 2015).

Third, software improvements can enable an agent to perform a task with the same level of proficiency on the same hardware system more quickly, and vice versa. For fixed hardware, a more efficient algorithm enables faster performance, and for fixed software, faster hardware enables faster performance. Borrowing Carroll's (1993) terminology for human intelligence differentiation, hardware and software can both contribute to increasing the *level* (discussed in the first point above) as well as the *speed* of AI performance.

Finally, beyond the level or speed of AI performance, hardware progress can accelerate algorithmic progress in AI indirectly by allowing researchers to run more (and a wider range of) experiments in a given amount of time. Thus, insofar as the AI community is conceived as engaged in a search for algorithms capable of faster, higher level performance for a given unit of hardware, hardware progress accelerates that search above and beyond its effect on individual system performances.

These four connections between hardware and software progress suggest some building blocks of a possible model of AI progress—namely, such a model should explicitly represent hardware progress as an independent variable, and future scenarios for AI progress should consider uncertainty in hardware developments. The four linkages discussed indicate ways that this model could be structured, but more work (discussed in "Future Directions" below) is needed to empirically evaluate different structures and parameters for such a model.

**Human Input**

In addition to hardware, the role of humans in facilitating AI performance should be considered in a rigorous model of AI progress. Specifically, humans fit into the picture of AI progress in at least two ways that must be accounted for.

First, the type and extent of human input into an algorithm's performance is inversely proportional to the AI progress demonstrated by that algorithm's performance. In other words, an AI demonstration is more impressive to the extent that there is little to no human fine-tuning of the AI system in order to achieve a given level of performance. This relationship comports well with common narratives related to AI progress– e.g. the impressiveness of the human-level control shown by DeepMind's DQN algorithm on a wide range of Atari games without requiring human adjustment of the hyperparameters for each game, or any human assistance in identifying the games' visual features in the input stream (Mnih et al., 2015). One could also explicitly consider the quality of human input in accounting for agent performance, in addition to the quantity, so as to model the progress involved in systems becoming usable by non-expert users. A given level of AI performance, in this formulation, represents more progress if it is attainable by a non-expert than if it required the labor of a graduate student to apply the algorithm to a particular domain. Importantly, this analysis is not intended to suggest that human interaction with an agent is a bad thing from the perspective of AI progress, all things considered—indeed, greater transparency, interactivity, and human-comprehensibility of AI systems can be beneficial from a societal perspective, and enabling them requires algorithmic innovations. The point here is simply that, other things being equal, less or lower-quality human input for a given performance level or speed output from an AI system represents a certain form of (performance-oriented) progress. Additionally, greater human input can coincide with greater (performance-oriented) AI progress when it is used to enable the human-machine system to perform at a higher level or with greater speed, so the ceteris paribus assumption is critical.

Second, humans are (for the foreseeable future, at least – cf. Bostrom, 2014) causally responsible for AI progress, as they design the algorithms in question. In the search for faster, higher level system performance, AI researchers advance the state(s) of the art, however defined. Anticipating AI progress, then, should involve some sort of assessment of the scale and quality of this human effort at algorithmic innovation, represented by a metric such as quality-adjusted research years (Muehlhauser and Sinick, 2014). In technological progress modeling work for other domains, such as clean energy (Bettencourt et al., 2013), metrics like funding have been used to approximate the effort applied to technology improvement. As suggested in "Future Directions" below, such measures could be compared with historical rates of progress in AI in order to discern causal dynamics.

**AI Sub-Fields and System Architecture**

A final area we will explore in order to help lay the foundation for rigorous AI progress modeling is the relationships between different AI sub-fields and integrated system architectures. The motivation for this section is the common claim that progress in the sub-fields of AI has been rapid, whereas progress on generally intelligent integrated systems has been more limited (Dietterich and Horvitz, 2015). In recent years, some benchmarks have been put forward to evaluate the general intelligence of AI systems (e.g. Bellemare et al., 2012). However, even with such benchmarks available, it is not obvious how to model the relationship of progress in AI sub-fields to progress on general intelligence, or how to model progress in developing integrated agent architectures. This sub-section suggests some possible ways to conceptualize these relationships.

As previously noted, Hernandez-Orallo (2014) suggests a possible mapping between "abilities" in a commonly used hierarchical model of intelligence (Carroll, 1993) and sub-fields of AI. However, there is not a perfect mapping between these, since the contours of human and machine intelligence are currently distinct in various ways, and the sub-field topology of AI also represents the cumulative effects of myriad idiosyncratic human decisions over several decades. One possible approach, then, is to develop, drawing on the literatures outlined above, an empirically and theoretically informed list of abilities, which may or may not map directly to AI sub-fields, which can be used to extrapolate future capabilities for different agent components. For example, sub-field/ability clusters could include natural language processing, machine learning, social interaction, perception, high level cognition, and other areas, which in turn could be broken into smaller components, with different metrics and models for evaluating each. Since these sub-fields/abilities are not independent (machine learning, for example, can assist natural language processing), more work would need to be done to develop principled groupings of abilities and connections between them. A further question that would then arise is: how could such a model be used to anticipate the ease of developing integrated systems that draw upon those abilities?

One way to address this could come from the study of human intelligence (van der Maas et al., 2006), where researchers have modeled positive interactions between cognitive abilities in order to explain the phenomenon of general intelligence and the correlation between highly intelligent humans' abilities. In this approach, the integration of capabilities would be represented by positive (signifying the synergistic effects of combining multiple abilities) and/or negative interactions (signifying the difficulty of integration without loss to capability) between cognitive abilities in a single system. One could then extrapolate progress in specific abilities as well as the future ease of developing agents with a particular combination of cognitive abilities. Further theoretical and empirical analysis is needed to develop a rigorous approach to such issues, and as previously noted, different approaches are suitable for different research questions. In the next section, we examine what sorts of modeling approaches may be appropriate to studying the economic implications of AI in particular.

**AI Progress and the Future of Work**

Prior work has explored the potentially enormous impact of AI on the economy (Brynjolfsson and McAfee, 2014; Vardi, 2015). Yet considerable uncertainty remains about the precise nature and timing of that impact, considering the many technical and social uncertainties. Different theories of how it is to automate different tasks (and, relatedly, different models of AI progress) imply different conclusions about the type and magnitude of those impacts. Thus, in this section, we survey these divergent theories of auto-mation difficulty and suggest approaches to AI progress modeling that could yield novel insights into the future of work.

**Theories of Automation Difficulty**

Several theories relating the susceptibility of jobs to automation and the state of the art in AI have been put forward. Murnane and Levy (2004) distinguish between routine and non-routine tasks, with the former being more susceptible to automation. Frey and Osborne (2013) find social intelligence, creative intelligence, and perception and manipulation to be "bottlenecks" to automation of particular tasks over the next two decades. Autor (2013) notes that novel tasks in the workplace are less susceptible to automation than long-established ones. Brynjolfsson and McAfee (2014) focus on creativity as a distinguishing factor between future jobs' degrees of vulnerability, while noting the considerable uncertainty in any such assessment. Rus (2015) suggests that different types of perception and manipulation tasks are suitable for humans versus machines, contra Frey and Osborne's (2013) more categorical distinction, and she also adds abstraction and creativity to the list of hard-to-automate abilities. Finally, Deming (2015) finds empirically that jobs drawing more on social intelligence have been safer in recent decades.

These different theories of automation difficulty have different implications for which jobs are likely to be automated (and, relatedly, which jobs are likely to be created as a result of complementarities between human and machine capabilities—Brynjolfsson and McAfee, 2014). Further development of AI progress models will be needed to differentiate between these theories and to derive strong conclusions about future scenarios. Notably, each of the above models is static—it posits a particular state of the art, and doesn't allow for change over the coming decades. Additionally, none takes into account issues such as architecture and integration, hardware progress, or many of the other issues raised in earlier sections. Thus, there is a need for improved analysis of automation scenarios using robust AI progress models. However, different modeling approaches are appropriate to analyzing the issue of technological unemployment versus, say, intelligence explosion. Thus, in the next sub-section, we consider which modeling approaches are suitable for anticipating the future of work.

**Tailored Progress Modeling Approaches**

Discerning trends in AI progress that are relevant to the future of work raises challenges that are different than those found in evaluating AI from certain scientific perspectives. For example, Legg and Hutter (2007) propose a universal intelligence measure that is rigorously defined, mathematically elegant, and well-justified, but is not necessarily ideal for thinking about economic implications of AI. Specifically, it favors generality of agents across the space of possible environments, even if those environments aren't economically (or even humanly) relevant. An agent may be economically profitable and widely applicable even if it is only adapted to a narrow slice of the space of possible environments. Additionally, modeling progress in AI in order to inform economic analysis requires considering the different levels of performance that may be called for in the economy, which may differ from the levels of performance (e.g. asymptotic optimality) demanded by AI researchers. Thus, economically-relevant AI progress modeling requires considering the demand for, and not just the supply of, intelligence—i.e. what sorts of agents consumers are willing to be served by in an economic context, at what price, with what levels of robustness in terms of the agent's skills, etc. This may call for developing new metrics of progress, as well as empirically evaluating (through, e.g. surveys or experiments) consumer demand for different levels of human/agent capabilities in the marketplace.

In addition to considering demand, researchers examining the susceptibility of tasks to automation should seriously consider the human input required for a given level of AI performance, as discussed earlier. Even if it is possible in principle to build a system with human-level or beyond performance in arbitrary domains, this may not be particularly economically significant if doing so requires advanced expertise and significant expenditures of time by humans in order to adapt the system to a new domain each time. Additionally, factors like robustness and worst-case performance may be especially important in economic contexts, to an extent that exceeds the requirements of scientific research oriented toward benchmarks such as those of Legg and Hutter (2007). An AI system may exceed human performance in almost all cases, but fail catastrophically in the remaining cases in a way that humans never would, making it economically undesirable for self-interested companies seeking to avoid lawsuits. This suggests that a rich array of metrics might be needed to evaluate economic feasibility of automating certain tasks.

Next, modeling AI progress for the purpose of economic foresight demands the use of metrics that can capture degrees of progress toward certain goals, rather than simply assessing in a binary fashion a goal's achievement or not, as with, e.g. the Turing Test. In the case of creativity, for example, a test like Riedl's (2014) Lovelace Test 2.0, which admits degrees of achievement, would be more useful than the original Lovelace Test (Bringsjord et al., 2001). Additionally, concepts like the Turing Ratio (Masum et al., 2003), which measures degrees of progress toward and beyond human-level performance, would assist in discerning trends toward minimal levels of performance that humans might accept in the marketplace, or toward superhuman levels of speed, performance, or robustness that might result in a strong consumer preference for being served by AIs over humans.

Another factor of particular relevance in economically-relevant analysis of AI is the cost of non-computing power-related hardware, such as robot sensor and effectors. Trends in the cost of robotic hardware may be especially relevant in determining the potential variation in the consequences of AI across space (due, e.g., to varying human wages). Such trends are, to some extent, already addressed in robotics roadmaps, but further research could explore the possible connections between AI performance levels and hardware platform availability.

Another potentially critical variable to consider in modeling automation difficulty is the availability of training data. As with increased hardware performance, more abundant training data can improve AI performance, and may be unevenly available across different economically relevant task domains.

Finally, as discussed in the sub-section "AI Sub-Fields and System Architecture," it is important to consider the architectural issues involved in developing complete AI systems, and this is especially true in the case of economic analysis. In analyzing the possibility of technological unemployment, Autor (2013) emphasizes the significance of "bundling" in labor markets—that is, cases where a human provides multiple services simultaneously (e.g. delivering food as well as socially engaging a customer). The frequency of such bundling suggests that analysts of technological unemployment should be particularly attuned to the issues of integration and system architecture discussed above.

# Future Directions

The analysis above suggests the need for additional research on a variety of different aspects of AI progress. Below, we group some directions for future research by whether they are primarily conceptual, empirical, or normative in nature, though there is overlap between these.

## Conceptual

The section above, "Toward Rigorous Modeling of AI Progress," suggested some initial ways to conceptualize and, eventually, quantitatively model AI progress, but it did not integrate the full range of relevant literatures. Unanswered questions include: what insights can be gained from the literature on natural intelligence in thinking about how (not) to evaluate AI progress, and in particular the recalcitrance (Bostrom, 2014) of various research problems? Are different models of AI progress needed for conceptualizing AI in the near term versus in the longer term during a future intelligence explosion (or lack thereof)? And are there more useful ways of representing the relationships between sub-fields of AI and the challenge of building integrated AI systems than those discussed above?

## Empirical

The analysis above raised a number of questions that require empirical investigation in order to answer. In each dimension of AI progress discussed, there is the empirical question of what the rate of progress has been historically and how it is linked to the posited independent variables (research effort, hardware acceleration, data availability, etc.). Fleshing out the models suggested here would require gathering additional data on such factors, and in many cases, quantifying elements of progress that so far have only been stated qualitatively in the literature. Also, developing a rich map of current research efforts across AI in order to extrapolate rates of progress in various sub-fields would require additional data gathering and analysis, through e.g. bibliometric investigation and analysis of funding trends. Finally, with regard to the economic dimensions of AI, questions were raised about consumer demand for different human/agent capabilities that would need to be investigated empirically in order to inform appropriate metrics for progress modeling.

## Normative

Anticipating plausible developments in AI and their implications, while difficult, is part of the ethical responsibility of the AI community (Brundage, forthcoming). The issues discussed in this paper raise various questions about such responsibilities—should, for example, progress in some areas be accelerated, decelerated, or modulated in some other way? Outside of the AI community, what responsibilities are there for corporations, policy-makers, consumers, and others to influence rates/types of AI progress? Should, as some suggest (Russell, 2014), the aims of the field of AI be redefined in a way that reflects the need for systems to be provably aligned with human values? If so, what metrics would be appropriate for capturing that dimension of AI progress?

# Conclusion

This paper has defended a research program aimed at rigorously modeling progress in AI, and has made some preliminary contributions to that program. Existing efforts to account for AI progress were found to be limited in important respects, and controversy in the literature was found with respect to the economic implications of AI, suggesting the need for further research. Approaches to thinking about and, ultimately, quantitatively modeling different aspects of AI progress were outlined in areas such as the software/hardware nexus, the role of human input, and the relationships between AI sub-fields. Finally, re-search directions were outlined that could shed further light on the nature, pace, and drivers of AI progress and their normative dimensions. Overall, AI progress modeling appears to be an area ripe for further investigation and one with significant social urgency.

## Acknowledgments

The author would like to acknowledge the helpful comments of David Guston, Joanna Bryson, Erik Fisher, Mark Gubrud, Daniel Dewey, Katja Grace, Kaj Sotala, Brad Knox, Vincent Mueller, Beau Cronin, Adi Wiezel, David Dalrymple, Adam Elkus, an anonymous reviewer, and others who commented on earlier versions of these ideas.